\documentclass[conference,10pt]{IEEEtran}
\IEEEoverridecommandlockouts

\usepackage{makecell}
\usepackage{float}
\usepackage{amsmath,amssymb,amsfonts}
\usepackage{graphicx}
\usepackage{textcomp}
\usepackage{xcolor}
\usepackage{newtxtext}
\usepackage{caption}
\usepackage{newtxmath}
\usepackage{setspace}
\usepackage{soul}
\usepackage{algorithm}
\usepackage{algpseudocode}
\usepackage{booktabs}
\usepackage[dvipsnames]{xcolor}
\usepackage[backend=biber, style=numeric, sorting=none]{biblatex} 
\algnewcommand\algorithmicinput{\textbf{Input:}}
\algnewcommand\algorithmicoutput{\textbf{Output:}}
\algnewcommand\Input{\item[\algorithmicinput]}
\algnewcommand\Output{\item[\algorithmicoutput]}
\def\BibTeX{{\rm B\kern-.05em{\sc i\kern-.025em b}\kern-.08em
T\kern-.1667em\lower.7ex\hbox{E}\kern-.125emX}}
\begin{document}

\title{GEGO: A Hybrid Golden Eagle and Genetic Optimization Algorithm for Efficient Hyperparameter Tuning in Resource-Constrained Environments}

\author{\IEEEauthorblockN{Amaras Nazarians, Sachin Kumar${^*}$}
\IEEEauthorblockA{Zaven P. and Sonia Akian College of Science and Engineering, American University of Armenia, Yerevan, Armenia} \\

amaras\_nazarians@edu.aua.am; s.kumar@aua.am \\ Corresponding author: Sachin Kumar(s.kumar@aua.am)}

\maketitle

\begin{abstract}
Hyperparameter tuning is a critical yet computationally expensive step in training neural networks, particularly when the search space is high dimensional and nonconvex. Metaheuristic optimization algorithms are often used for this purpose due to their derivative free nature and robustness against local optima. In this work, we propose Golden Eagle Genetic Optimization (GEGO), a hybrid metaheuristic that integrates the population movement strategy of Golden Eagle Optimization with the genetic operators of selection, crossover, and mutation.

The main novelty of GEGO lies in embedding genetic operators directly into the iterative search process of GEO, rather than applying them as a separate evolutionary stage. This design improves population diversity during search and reduces premature convergence while preserving the exploration behavior of GEO.

GEGO is evaluated on standard unimodal, multimodal, and composite benchmark functions from the CEC2017 suite, where it consistently outperforms its constituent algorithms and several classical metaheuristics in terms of solution quality and robustness. The algorithm is further applied to hyperparameter tuning of artificial neural networks on the MNIST dataset, where GEGO achieves improved classification accuracy and more stable convergence compared to GEO and GA. These results indicate that GEGO provides a balanced exploration–exploitation tradeoff and is well suited for hyperparameter optimization under constrained computational settings.
\end{abstract}


\section{Introduction and Background}
Optimization algorithms are widely used across many domains, including engineering, computer science, and business. In machine learning, optimization plays a central role in tasks such as minimizing model loss or maximizing prediction accuracy. In addition to classical analytical optimization techniques, metaheuristic algorithms have gained significant attention due to their derivative free nature and their ability to handle complex, nonconvex search spaces without strong assumptions.

Metaheuristic algorithms are commonly categorized into evolutionary based methods and swarm intelligence based methods. Evolutionary algorithms update a population of candidate solutions using fitness driven operators inspired by natural evolution \cite{Bozorg}. In contrast, swarm intelligence algorithms are motivated by the collective behavior of social animals, where individual agents use both personal experience and shared group information to guide the search process. One of the earliest and most well known swarm based algorithms is Particle Swarm Optimization (PSO). In PSO, each particle maintains a position and velocity, and the velocity is updated based on both the particle’s individual best position and the best position found by the group \cite{PSO}. Other swarm based algorithms adopt different movement strategies and interaction mechanisms to balance exploration and exploitation.

Among the large family of metaheuristic algorithms, Golden Eagle Optimization and Genetic Algorithms exhibit complementary search characteristics that motivate their hybridization. GEO provides directed population movement through its attack and cruise mechanisms, which supports effective global exploration of the search space. However, similar to many swarm based methods, population diversity may decrease in later iterations, increasing the risk of premature convergence. Genetic Algorithms, on the other hand, explicitly promote diversity through selection, crossover, and mutation, but often lack guided movement toward promising regions of the search space. By combining these two approaches, the objective of this work is to retain the exploration capability of GEO while using genetic operators to sustain diversity throughout the optimization process.

In this paper, we propose a new metaheuristic algorithm called Golden Eagle Genetic Optimization (GEGO). GEGO is a hybrid of Golden Eagle Optimization, a swarm intelligence algorithm, and the Genetic Algorithm, which belongs to the evolutionary computation family. The performance of GEGO is evaluated using standard mathematical benchmark functions and is further demonstrated through its application to hyperparameter tuning of artificial neural network models.

\subsection{Contribution of the Work}
The contribution of this work is not the introduction of a completely new search principle, but the design of a tightly integrated hybrid optimization framework. Unlike existing GEO variants or loosely coupled GEO GA hybrids, GEGO embeds crossover and mutation operations within the iterative position update process of GEO at predefined intervals. This allows diversity enhancement without disrupting the swarm intelligence dynamics of GEO. To the best of our knowledge, GEO has not previously been applied to neural network hyperparameter tuning, and no prior study has investigated a unified GEO GA hybrid for this task under limited computational budgets.



\section{Related Work}
The Golden Eagle Optimization (GEO) algorithm and its variants have been applied to a wide range of optimization problems and have demonstrated competitive performance across different application domains. Eluri and Devarakonda \cite{ELURI2022108771} proposed a time varying GEO variant for feature selection and reported improved convergence behavior and classification accuracy. Similarly, Siva et al. \cite{siva_2023_automatic} employed GEO for feature selection within a deep learning framework for automatic software bug prediction, highlighting its effectiveness in handling high dimensional feature spaces.

Several studies have explored hybridization strategies involving GEO. Lv et al. \cite{lv_2022_a} combined GEO with the Grey Wolf Optimizer (GWO) \cite{GWO} to address three dimensional path planning for unmanned aerial vehicle based power line inspection, achieving improved exploration capability and solution quality. In the context of cloud computing, Jagadish Kumar and Balasubramanian \cite{jagadishkumar_2023_hybrid} integrated GEO with gradient descent to optimize heterogeneous resource scheduling for big data processing, resulting in reduced execution time and improved scheduling efficiency.

Existing applications of GEO in neural network optimization have primarily focused on weight optimization rather than model configuration. Amor et al. \cite{amor_2022_comfort} incorporated GEO into the ANN learning process to optimize network weights for comfort evaluation of coated fabrics and reported improved predictive performance. However, hyperparameter tuning was not considered in their study, despite representing a more complex and computationally demanding optimization task.

In contrast to GEO based approaches, several studies have investigated hyperparameter optimization using other metaheuristic algorithms. Gaspar et al. \cite{Gaspar2021} evaluated multiple metaheuristic methods for convolutional neural network hyperparameter optimization on the MNIST dataset, although their comparison did not include Genetic Algorithms or GEO. Similarly, Nematzadeh et al. \cite{NEMATZADEH2022107619} applied Genetic Algorithm based hyperparameter tuning to machine learning and deep learning models in bioinformatics, emphasizing the effectiveness of evolutionary search strategies for complex learning problems.

Despite the growing body of work on GEO and metaheuristic based hyperparameter optimization, there is currently no reported implementation of GEO for neural network hyperparameter tuning. Furthermore, a hybrid framework that integrates GEO with Genetic Algorithms in a unified optimization process has not been explored in the existing literature. This gap motivates the proposed approach.

Recent trends in metaheuristic optimization research increasingly emphasize hybrid and adaptive frameworks that aim to combine complementary search behaviors. At the same time, there is growing recognition of the importance of computational efficiency and resource awareness, particularly in learning-based applications where fitness evaluation is expensive. While large-scale benchmarks and extensive comparisons are valuable, controlled studies under constrained settings remain relevant for understanding algorithmic behavior and robustness. In this context, the present work contributes to the literature by examining a unified hybrid design that emphasizes integration efficiency rather than scale alone.

\subsection{Summary of Existing GEO-Based and Metaheuristic Approaches}

Table~\ref{tab:related_work_summary} summarizes representative studies related to GEO, its hybrid variants, and metaheuristic based hyperparameter optimization. The table highlights the application domains, key strengths, and limitations of existing approaches, and clarifies the research gap addressed in this work.

\begin{table*}[htbp]
\centering
\caption{Summary of related GEO-based and metaheuristic optimization studies}
\label{tab:related_work_summary}
\begin{tabular}{p{2.5cm} p{3cm} p{3cm} p{3cm} p{3cm}}
\hline
\textbf{Reference} & \textbf{Optimization Method} & \textbf{Application Domain} & \textbf{Key Advantage} & \textbf{Limitation} \\
\hline
Eluri and Devarakonda \cite{ELURI2022108771} & Time-varying GEO & Feature selection & Improved convergence and accuracy & Not applied to neural networks \\
Siva et al. \cite{siva_2023_automatic} & GEO with deep learning & Software bug prediction & Effective handling of high dimensional features & Focused on feature selection only \\
Lv et al. \cite{lv_2022_a} & GEO + GWO hybrid & UAV path planning & Enhanced exploration capability & Application specific evaluation \\
Jagadish Kumar and Balasubramanian \cite{jagadishkumar_2023_hybrid} & GEO + gradient descent & Cloud resource scheduling & Reduced execution time & Not evaluated for learning models \\
Amor et al. \cite{amor_2022_comfort} & GEO-assisted ANN & ANN weight optimization & Improved predictive performance & No hyperparameter tuning \\
Gaspar et al. \cite{Gaspar2021} & Multiple metaheuristics & CNN hyperparameter tuning & Comparative analysis of optimizers & GEO and GA not included \\
Nematzadeh et al. \cite{NEMATZADEH2022107619} & Genetic Algorithm & Hyperparameter tuning & Effective evolutionary search & No swarm-based methods considered \\

\hline
\end{tabular}
\end{table*}

While several recent optimization algorithms and large scale evaluation benchmarks have been proposed in the literature, comprehensive experimental comparisons with all modern optimizers require substantial computational resources. The experimental scope of this study was therefore designed to allow a controlled and reproducible evaluation under limited computational settings. Future work with access to larger computational budgets can extend the comparison to additional recent optimizers and more complex datasets.

\subsection{Recent Advances in Hyperparameter Optimization}

In recent years, hyperparameter optimization has received significant attention, leading to the development of advanced and increasingly specialized optimization frameworks. Notable examples include Bayesian optimization based methods, population based training, and large scale evolutionary strategies. Bayesian optimization techniques such as Tree structured Parzen Estimators and Gaussian process based methods have demonstrated strong performance, particularly for low dimensional and moderately expensive objective functions. More recently, scalable variants and hybrid approaches have been proposed to address higher dimensional search spaces and complex learning models.

In the evolutionary optimization literature, state of the art optimizers such as LSHADE and its variants have achieved strong results on large scale continuous benchmark suites, including CEC competitions. These methods rely on sophisticated parameter adaptation mechanisms and often require large function evaluation budgets to fully realize their performance advantages. Similarly, neural architecture search frameworks and gradient based hyperparameter optimization methods have gained popularity in deep learning applications, although they typically involve substantially higher computational costs and more complex implementation pipelines.

While these advanced methods represent important developments in the field, their evaluation often assumes access to extensive computational resources and large scale experimental settings. The present study focuses instead on a controlled and resource constrained evaluation scenario, with the objective of examining the effectiveness of a tightly integrated hybrid metaheuristic under limited population sizes and iteration budgets. Accordingly, the experimental comparisons emphasize representative and widely used population based optimizers that are commonly applied in practical hyperparameter tuning scenarios. A more comprehensive comparison with recent large scale and state of the art optimizers is identified as an important direction for future work.

\section{Methodology}
\subsection{Golden Eagle Optimization (GEO)}
The Golden Eagle Optimization (GEO) algorithm is a swarm intelligence based metaheuristic inspired by the hunting behavior of golden eagles. In GEO, each search agent represents a candidate solution and updates its position using a combination of directed movement toward a selected prey and lateral exploration around that direction. This section summarizes the core principles of GEO that are relevant to the proposed hybrid framework.

\begin{algorithm}[]
\caption{GEO Algorithm}
\label{alg:geo}
\begin{algorithmic}[1]
\Input{Fitness function = f, population size = N, Number of Iteration = T}
\Output{$gbest,X_{gbest}$}
\State Initialize the population of golden eagles
\State Initialize population memory
\State Initialize $p_a$ and $p_c$
\For{$t$ from 1 to $T$}
    \State Update $p_a$ and $p_c$ (Equation (5)):
    \For{each golden eagle $i$}
        \State Randomly select a prey from the population's memory
        \State Calculate \textbf{attack vector} $\vec{A}_i$ (Equation (1))
        \If{$\lVert\vec{A}_i\rVert \neq 0$}
            \State Calculate \textbf{cruise vector} $\vec{C}_i$ (Equation (3))
            \State Calculate \textbf{step vector} $\Delta x_i$ (Equation (4))
            \State Update position $x_i$  (Equation (6)) 
            \State  $y_i \gets  f(x_i^{t+1})$
            \If{ $y_i \le pbest_i$}
                \State Replace the position in eagle $i$'s memory with $x_i^{t+1}$
                \State $pbest_i\gets y_i$
            \EndIf
        \EndIf
    \EndFor
\EndFor
\State $gbest \gets \min _{\forall \text{golden eagle i}} \{ pbest_i \}$\\
\Return{gbest}
\end{algorithmic}
\end{algorithm}
This model is inspired by the hunting behavior of Golden Eagles. Each particle maintains an \textit{attack vector} and a \textit{cruise vector}, which collectively determine its trajectory toward the prey. The attack vector represents the displacement from the particle's current position to the prey's location. The prey is selected randomly from the population's memory; specifically, rather than using a particle's own historical best, a particle is chosen at random from the population, and its best-found position is utilized. The attack vector $\vec{A}_i$ is defined in Equation \ref{eq:attack}, where $\vec{X}_f^*$ denotes the selected best position.

\begin{equation}
\vec{A}_i = \vec{X}_f^* - \vec{X}_i
\label{eq:attack}
\end{equation}

The cruise vector $\vec{C}_i$ is calculated as a tangent vector to the circle and is perpendicular to the attack vector. In an $n$-dimensional search space, the cruise vector resides within a hyperplane where the attack vector serves as the normal vector. To compute $\vec{C}_i$ in $n$ dimensions, $n-1$ components are assigned random values, while the final variable is determined by the hyperplane equation:
\begin{equation}
\sum_{j=1}^{n} a_j x_j = \sum_{j=1}^{n} a_j^{t} x_j^{*}
\end{equation}

The $k$-th component of the cruise vector is then derived as:
\begin{equation}
c_k = \frac{d - \sum_{j \neq k} a_j c_j}{a_k}
\end{equation}
where the index $k$ is selected such that $a_k \neq 0$. The total displacement of the particle is defined by Equation \ref{eq:displacement}:
\begin{equation}
\Delta x_i = r_1 p_a \frac{\vec{A}_i}{\lVert\vec{A}_i\rVert} + r_2 p_c \frac{\vec{C}_i}{\lVert\vec{C}_i\rVert}
\label{eq:displacement}
\end{equation}
Here,  $r_1$ and $r_2$ are random numbers from the interval [0,1] and $p_a$ and $p_c$ are the attack and cruise coefficients, respectively. These coefficients are time-variant, adjusting linearly over iterations $t$ relative to the maximum iterations $T$ as follows:
\begin{equation}
\begin{cases}
p_a = p_a^0 + \frac{t}{T}|p_a^T - p_a^0| \\
p_c = p_c^0 - \frac{t}{T}|p_c^T - p_c^0|
\end{cases}
\end{equation}
Algorithm 1, represents the optimization algorithm for a given function for minimization. As a input we define the population size and the maximum number of iterations. The ending condition is based on the maximum number of iterations, where in each iteration we update the position of all particles according to Equation \ref{eq:update}. 

\begin{equation}
x^{t+1} = x^t + \Delta x_i^t
\label{eq:update}
\end{equation}
\subsection{Genetic Algorithm (GA)}
The Genetic Algorithm (GA) is an evolutionary computation method consisting of three primary stages: selection, crossover, and mutation. In GA, candidate solutions are encoded as chromosomes, often in binary format. During the crossover stage, a fraction of the population is selected to generate offspring through genetic recombination. These offspring then undergo random mutation based on a predefined mutation rate. If an offspring demonstrates superior fitness compared to its parent, the parent is replaced in the next generation.
\begin{figure}[h]
    \centering
    \includegraphics[width=0.5\textwidth]{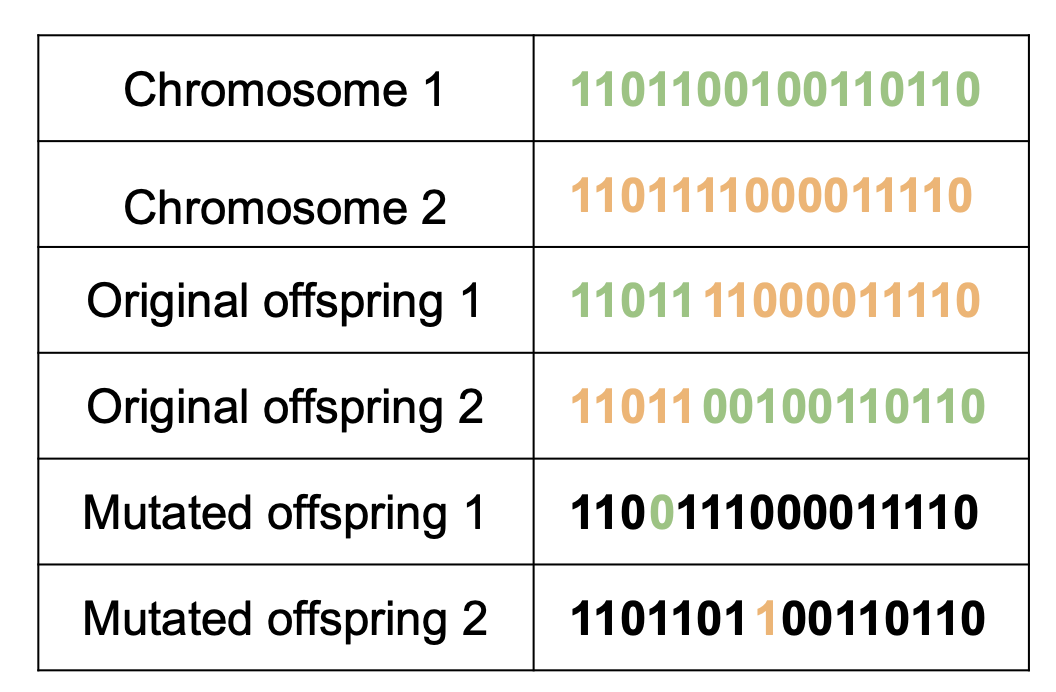}
    \caption{Process for a pair of 16-bit chromosomes}
    \label{fig:fig1}
\end{figure}
Figure \ref{fig:fig1} illustrates this process for a pair of 16-bit chromosomes. In this example, a crossover point at the eleventh bit is chosen; Offspring 1 inherits the five leftmost bits from Parent 1 and the remaining eleven from Parent 2. Following crossover, mutation is applied specifically to the fourth and eighth bits in this instance. 
If mutated offspring 1 has a better fitness value compared to parent 1, i.e. function at the mutated offspring is lower than at chromosome 1, then parent 1 gets replaced by mutated offspring 1. Same comparison is done for the parent 2 and the mutated offspring 2. 

\subsection{Golden Eagle Genetic Optimization (GEGO)}

This work proposes a hybrid metaheuristic framework named Golden Eagle Genetic Optimization (GEGO), which integrates genetic operators into the core search process of the Golden Eagle Optimization (GEO) algorithm. The motivation behind GEGO is to address the gradual loss of population diversity that can occur in swarm based algorithms during later iterations, while preserving the directed movement and exploration capabilities of GEO.

In GEGO, each search agent follows the standard GEO position update mechanism based on attack and cruise vectors. In addition to this movement strategy, genetic operators are periodically applied to the population at predefined iteration intervals. Specifically, after a fixed number of GEO iterations, the continuous position vectors of all particles are temporarily encoded into chromosome representations. This encoding enables the application of crossover and mutation operations in a manner similar to classical Genetic Algorithms.

Unlike traditional GA approaches that typically apply genetic operators only to a selected subset of elite individuals, GEGO applies crossover and mutation to the entire population. This design choice is intended to maintain global diversity and reduce the risk of premature convergence, particularly in complex or multimodal search spaces. Following the genetic operations, the resulting chromosomes are decoded back into continuous particle representations. If the newly generated particle exhibits improved fitness compared to its parent, it replaces the parent in the population memory.

\begin{figure*}[]
    \centering
    \includegraphics[width=0.8\textwidth]{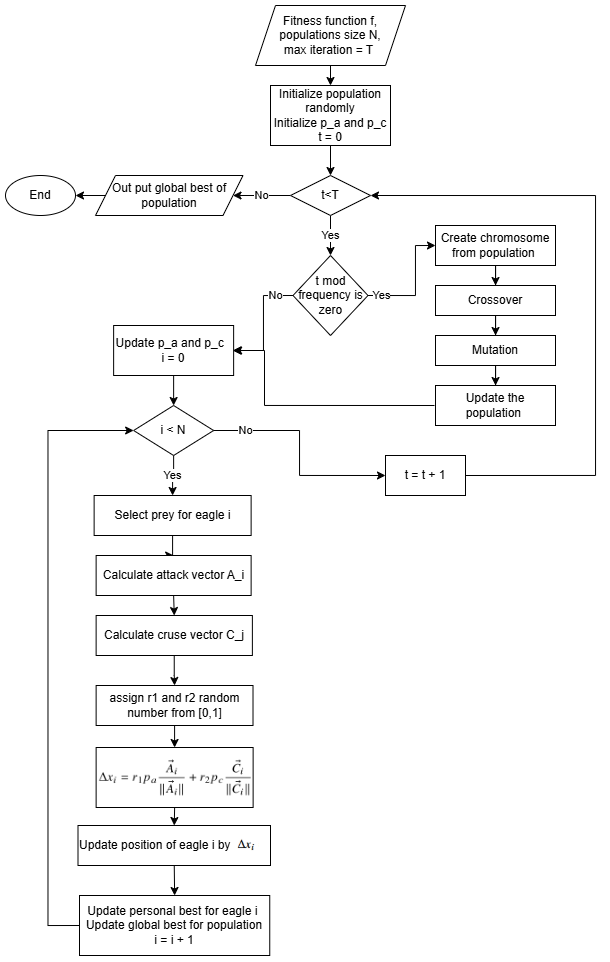}
    \caption{Graphical workflow of GEGO algorithm}
    \label{fig:gego}
\end{figure*}

The genetic phase does not replace the GEO dynamics but complements them. After the genetic update, the algorithm resumes the standard GEO position update process using the updated population. By embedding genetic operators directly within the iterative GEO framework, GEGO achieves a balanced interaction between swarm intelligence driven exploration and evolutionary diversity preservation. Figure \ref{fig:gego} shows the graphical workflow of the GEGO algorithm.


Algorithm 2 summarizes the complete GEGO procedure. The frequency of genetic intervention, mutation rate, and crossover strategy are treated as hyperparameters and are kept consistent across all experiments to ensure fairness. This integrated design allows GEGO to enhance search robustness while maintaining computational efficiency, making it suitable for optimization tasks under limited computational budgets. To further analyze the efficiency of the proposed approach, the computational complexity of GEGO is discussed in the following subsection.

\begin{algorithm}[]
\caption{GEGO Algorithm}
\label{alg:geo}
\begin{algorithmic}[1]
\Input{Fitness function = f, population size = N, Number of Iteration = T}
\Output{$gbest,X_{gbest}$}
\State Initialize the population of golden eagles
\State Initialize population memory
\State Initialize $p_a$ and $p_c$
\For{$t$ from 1 to $T$}
    \If{t is divisible by frequency}
        \State Convert particles to N bit choromosem
        \State Choose random number k from 1 to N
        \State Do Crossover from the first k bits.
        \State Mutate the new chromosomes by  mutation\_rate
        \State reverse the chromosomes to the particles
        \If{new particles are better then parent}
            \State replace the parent
        \EndIf
    \EndIf
    \State Update $p_a$ and $p_c$ (Equation (5)):
    \For{each golden eagle $i$} 
        \State Randomly select a prey from the population's memory
        \State Calculate \textbf{attack vector} $\vec{A}_i$ (Equation (1))
        \If{$\lVert\vec{A}_i\rVert \neq 0$}
            \State Calculate \textbf{cruise vector} $\vec{C}_i$ (Equation (3))
            \State Calculate \textbf{step vector} $\Delta x_i$ (Equation (4))
            \State Update position $x_i$  (Equation (6)) 
            \State  $y_i \gets  f(x_i^{t+1})$
            \If{ $y_i \le pbest_i$}
                \State Replace the position in eagle $i$'s memory with $x_i^{t+1}$
                \State {$pbest_i\gets y_i$}
            \EndIf
        \EndIf
    \EndFor
\EndFor
\State $gbest \gets \min _{\forall \text{golden eagle i}} \{ pbest_i \}$\\
\Return{gbest}
\end{algorithmic}
\end{algorithm}

The design of GEGO is motivated by the need to balance directed search behavior with sustained population diversity throughout the optimization process. In the original GEO framework, the attack and cruise mechanisms provide effective global exploration by guiding agents toward promising regions of the search space. However, as iterations progress, the population may gradually lose diversity, which can lead to premature convergence. To address this issue without altering the fundamental dynamics of GEO, genetic operators are incorporated in a controlled and periodic manner. In GEGO, crossover and mutation are not applied continuously at every iteration. Instead, they are embedded at predefined intervals within the GEO update cycle. This design choice ensures that the swarm intelligence behavior of GEO remains dominant, while genetic operators act as a diversity reinforcement mechanism when stagnation begins to emerge. Crossover facilitates information exchange between high-quality solutions, whereas mutation introduces controlled randomness that enables the search to escape local optima. By integrating these operations directly into the iterative process rather than as a separate evolutionary stage, GEGO maintains a unified optimization flow with minimal disruption to GEO’s search trajectory.

\subsubsection{Handling Mixed Continuous and Discrete Search Spaces in GEGO}

The hyperparameter optimization problem considered in this work involves a mixed search space composed of continuous, integer, and categorical variables. In the proposed GEGO framework, this mixed nature is handled through a representation and decoding strategy that preserves the original optimization dynamics while ensuring that all candidate solutions evaluated during the search correspond to valid neural network configurations.

Although Algorithm~2 uses the term chromosome when describing the application of genetic operators, each search agent in GEGO is internally maintained as a continuous valued vector throughout the optimization process. Both the Golden Eagle Optimization update mechanism and the genetic operators act directly on this continuous representation. Discrete and categorical hyperparameters are therefore not optimized in their native form but are obtained through a deterministic decoding step that is applied prior to fitness evaluation.

Architectural hyperparameters such as the number of hidden layers and the number of neurons in each layer are represented implicitly as continuous variables during optimization. These values are decoded by rounding to the nearest integer and subsequently constrained to remain within the predefined bounds of the search space. When the decoded number of layers is smaller than the maximum supported network depth, the neuron values associated with unused layers are ignored during model construction. This ensures architectural feasibility without introducing additional constraints into the optimization process.

Discrete and categorical hyperparameters including batch size and dropout rate are decoded by mapping the continuous values to the nearest admissible options from predefined sets. The learning rate is treated as a continuous variable and optimized in logarithmic scale, which enables efficient exploration across multiple orders of magnitude while maintaining numerical stability. Boundary handling is applied during decoding to prevent infeasible configurations from being passed to the fitness evaluation stage.

This encoding and decoding strategy is commonly adopted in metaheuristic based hyperparameter optimization and allows population based optimizers originally designed for continuous domains to operate effectively on mixed variable problems. Importantly, this mechanism is applied only at the representation and evaluation level and does not alter the execution flow or search behavior of GEGO. As a result, every fitness evaluation performed during the optimization corresponds to a valid and fully specified neural network model, ensuring consistency, reproducibility, and alignment with the experimental results reported in this study.

\subsection{Computational Complexity and Overhead Analysis}

The computational complexity of the proposed GEGO algorithm is comparable to that of standard population based metaheuristic methods. Let N denote the population size, T the maximum number of iterations, D the dimensionality of the optimization problem, and F the computational cost of evaluating the fitness function. Similar to GEO and the Genetic Algorithm, the overall time complexity of GEGO is given by $O(N \times T \times (F + D))$.

The dominant computational cost in GEGO arises from fitness evaluations, as each candidate solution must be evaluated at every iteration. The additional operations introduced by the genetic phase, including crossover and mutation, involve vector level computations whose cost scales linearly with the dimensionality D. Since these genetic operations are applied periodically and operate on the existing population, they do not change the asymptotic order of complexity.

In practical scenarios where the fitness evaluation cost F is constant or significantly smaller than the cost associated with model training, the complexity simplifies to $O(N \times T \times D)$. This behavior is consistent with other swarm intelligence and evolutionary optimization algorithms. For neural network hyperparameter tuning, the computational overhead of the optimizer itself is negligible compared to the cost of training and evaluating the neural network models.

Overall, the integration of genetic operators within the GEO framework increases only the constant factors of computation while preserving the same asymptotic complexity. This makes GEGO suitable for optimization tasks under limited computational resources, as demonstrated in the experimental evaluation.

In addition to time complexity, the memory requirements of GEGO are comparable to those of standard population-based metaheuristics. The algorithm maintains a population of N candidate solutions, each represented by a D-dimensional vector, resulting in a memory complexity of O(N × D). The genetic operations do not introduce additional population structures and operate directly on existing individuals, which avoids extra memory overhead.

From a practical perspective, the additional computational cost introduced by crossover and mutation affects only constant factors and does not alter scalability with respect to population size or problem dimensionality. In resource-constrained environments, such as those considered in this study, the dominant cost arises from fitness evaluations rather than optimizer mechanics. Consequently, GEGO remains computationally feasible under limited hardware settings while providing improved search robustness compared to its constituent algorithms.

\section{Experiment}
\subsection{Mathematical Benchmarking}
To evaluate the performance of the proposed GEGO algorithm, it was tested on a suite of unimodal and multimodal functions, as well as the ten composite benchmark functions from the CEC2017 competition. The specific characteristics of these functions are summarized in Table \ref{tab:cec2017_composite_labels}. Detailed descriptions of the CEC2017 functions are available in \cite{Van_Thieu_2024_Opfunu}, and their implementation was facilitated by the \textit{Opfunu} library \cite{thieu_nguyen_2020_3711682}.

\begin{table*}[]
\centering
\caption{Composite benchmark functions of CEC2017 competition ($D=30$, Bounds $[-100,100]^D$)} 
\label{tab:cec2017_composite_labels}
\renewcommand{\arraystretch}{1.5}
\small
\begin{tabular}{llcc}
\toprule
\textbf{Label} & \textbf{Component Functions ($f_i$)} & \textbf{Parameters} & $f^\ast$ \\
\midrule

\textbf{CF1} & 
$\begin{cases}
f_1: \text{SR Rosenbrock} \\
f_2: \text{SR High Conditioned Elliptic} \\
f_3: \text{SR Rastrigin}
\end{cases}$
&
$\begin{array}{l}
\sigma = [10, 20, 30] \\
\lambda = [1, 10^{-6}, 1] \\
\text{bias} = [0, 100, 200]
\end{array}$
& 2100
\\ \midrule

\textbf{CF2} & 
$\begin{cases}
f_1: \text{SR Rastrigin} \\
f_2: \text{SR Griewank} \\
f_3: \text{SR Modified Schwefel}
\end{cases}$
&
$\begin{array}{l}
\sigma = [10, 20, 30] \\
\lambda = [1, 10, 1] \\
\text{bias} = [0, 100, 200]
\end{array}$
& 2200
\\ \midrule

\textbf{CF3} & 
$\begin{cases}
f_1: \text{SR Rosenbrock} \\
f_2: \text{SR Ackley} \\
f_3: \text{SR Modified Schwefel} \\
f_4: \text{SR Rastrigin}
\end{cases}$
&
$\begin{array}{l}
\sigma = [10, 20, 30, 40] \\
\lambda = [1, 10, 1, 1] \\
\text{bias} = [0, 100, 200, 300]
\end{array}$
& 2300
\\ \midrule

\textbf{CF4} & 
$\begin{cases}
f_1: \text{SR Ackley} \\
f_2: \text{SR Rastrigin} \\
f_3: \text{SR High Conditioned Elliptic} \\
f_4: \text{SR Griewank}
\end{cases}$
&
$\begin{array}{l}
\sigma = [10, 20, 30, 40] \\
\lambda = [1, 1, 10^{-6}, 10] \\
\text{bias} = [0, 100, 200, 300]
\end{array}$
& 2400
\\ \midrule

\textbf{CF5} & 
$\begin{cases}
f_1: \text{SR Rastrigin}, f_2: \text{SR HappyCat} \\
f_3: \text{SR Ackley}, f_4: \text{SR Discus} \\
f_5: \text{SR Rosenbrock}
\end{cases}$
&
$\begin{array}{l}
\sigma = [10, 20, 30, 40, 50] \\
\lambda = [10, 1, 10, 1, 1] \\
\text{bias} = [0, \dots, 400]
\end{array}$
& 2500
\\ \midrule

\textbf{CF6} & 
$\begin{cases}
f_1: \text{SR Expanded Schaffer's F6} \\
f_2: \text{SR Schwefel}, f_3: \text{SR Griewank} \\
f_4: \text{SR Rosenbrock}, f_5: \text{SR Rastrigin}
\end{cases}$
&
$\begin{array}{l}
\sigma = [10, 20, 20, 30, 40] \\
\lambda = [1, 1, 10, 1, 1] \\
\text{bias} = [0, \dots, 400]
\end{array}$
& 2600
\\ \midrule

\textbf{CF7} & 
$\begin{cases}
f_1: \text{SR H.C. Elliptic}, f_2: \text{SR Rastrigin} \\
f_3: \text{SR HappyCat}, f_4: \text{SR Rosenbrock} \\
f_5: \text{SR Mod. Schwefel}, f_6: \text{SR Ackley}
\end{cases}$
&
$\begin{array}{l}
\sigma = [10, 10, 10, 20, 20, 20] \\
\lambda = [10^{-6}, 10, 1, 1, 1, 10] \\
\text{bias} = [0, \dots, 500]
\end{array}$
& 2700
\\ \midrule

\textbf{CF8} & 
$\begin{cases}
f_1: \text{SR Rastrigin}, f_2: \text{SR Griewank} \\
f_3: \text{SR H.C. Elliptic}, f_4: \text{SR HappyCat} \\
f_5: \text{SR Discus}, f_6: \text{SR Rosenbrock}
\end{cases}$
&
$\begin{array}{l}
\sigma = [10, 20, 30, 40, 50, 60] \\
\lambda = [1, 10, 10^{-6}, 1, 1, 1] \\
\text{bias} = [0, \dots, 500]
\end{array}$
& 2800
\\ \midrule

\textbf{CF9} & 
$\begin{cases}
f_1: \text{SR H.C. Elliptic} \\
f_2: \text{SR Rastrigin} \\
f_3: \text{SR Rosenbrock}
\end{cases}$
&
$\begin{array}{l}
\sigma = [10, 30, 50] \\
\lambda = [1, 1, 1] \\
\text{bias} = [0, 100, 200]
\end{array}$
& 2900
\\ \midrule

\textbf{CF10} & 
$\begin{cases}
f_1: \text{SR H.C. Elliptic} \\
f_2: \text{SR Rastrigin} \\
f_3: \text{SR Rosenbrock}
\end{cases}$
&
$\begin{array}{l}
\sigma = [10, 30, 50] \\
\lambda = [0.1, 1, 10] \\
\text{bias} = [0, 100, 200]
\end{array}$
& 3000
\\
\bottomrule
\end{tabular}
\end{table*}

The performance of GEGO is compared against several well-established metaheuristic algorithms: Grey Wolf Optimizer (GWO) \cite{GWO}, Jellyfish Search Algorithm (JSA) \cite{JSA}, Sine Cosine Algorithm (SCA) \cite{SCA} Particle Swarm Optimization (PSO) \cite{PSO}, and its constituent algorithms, Golden Eagle Optimization (GEO) \cite{GEO} and Genetic Algorithm (GA) \cite{GA}. All algorithms were implemented in Python and executed on a 14-core CPU (M4 generation). The general and solver-specific parameters for each algorithm are reported in Table \ref{tab:parameter_settings}.

\begin{table}[ht]
\centering
\caption{Parameter settings for compared algorithms}
\label{tab:parameter_settings}
\resizebox{\columnwidth}{!}{
\begin{tabular}{lll}
\toprule
Algorithm & Parameter & Value \\
\midrule
GEGO & $p_a$: Propensity to attack & $[0.5 - 2]$ \\
    & $p_c$: Propensity to cruise & $[1 - 0.5]$ \\
    & Selection method & Binary tournament \\
    & Crossover method & Linear \\
    & Mutation probability & 0.001 \\
    & Frequency & 3 or 5 \\
\midrule
GEO & $p_a$: Propensity to attack & $[0.5 - 2]$ \\
    & $p_c$: Propensity to cruise & $[1 - 0.5]$ \\
\midrule
GA  & Elite fraction & 0.05 \\
    & Selection method & Binary tournament \\
    & Crossover method & Linear \\
    & Mutation probability & 0.001 \\
\midrule
GWO & $C$: Control parameter & $[2 - 0]$ \\
    & Number of leaders & 3 \\
\midrule
PSO & $w$: Inertia weight & 0.8 \\
    & $c_1, c_2$: Acceleration weights & 0.5 \\
\midrule
JSA &$\eta:$ Motion Constant& 4\\
    &$\beta:$ Distribution coefficient&3\\
    &$\gamma:$ Motion direction coefficient& 0.1\\
    &$c_0:$ Control constant& 0.5\\
\midrule
SCA &  linear\_component & 2\\
\midrule
L\_SHADE & $M_{CR}:$ Control constant CR & 0.5\\
        & $M_F:$ Control constant F & 0.5\\
\bottomrule
\end{tabular}
}
\end{table}

Since metaheuristic algorithms are stochastic in nature, their results can vary between runs due to random initialization and intermediate calculations. To ensure statistical significance and mitigate the impact of randomness, each algorithm was executed for 40 independent trials per benchmark problem. We report the resulting mean and standard deviation for each case. The reported results are presented in terms of mean and standard deviation over multiple independent runs, which is a common practice in the evaluation of stochastic metaheuristic algorithms. Due to the limited number of runs conducted under constrained computational resources, formal statistical significance tests such as Wilcoxon or Friedman tests were not applied. As a result, the performance comparisons should be interpreted as indicative rather than conclusive. Nevertheless, the consistent trends observed across benchmark functions and application tasks suggest the robustness of the proposed approach.

Table \ref{tab:math_simple_20} presents the results for the unimodal and multimodal functions. The results indicate that GEO outperforms GEGO in only two instances. In all other functions, GEGO either reaches the same global minimum as GEO or demonstrates GEGO exhibits superior performance compared to the considered baseline optimizers under the evaluated experimental settings.

\begin{table*}[]
\centering
\caption{Results of benchmark functions (max\_it=100, pop\_size=20)}
\label{tab:math_simple_20}
\small
\setlength{\tabcolsep}{4pt} 
\begin{tabular}{llccccccc}
\toprule
Function & & GEGO & GEO & PSO & GA & GWO & JSA & SCA \\
\midrule
Beale & Mean & $5.16\times10^{-4}$ & $6.42\times10^{-6}$ & $7.75\times10^{-2}$ & $9.51\times10^{-1}$ & $2.19\times10^{-6}$ & $1.75\times10^{-15}$ & $2.31\times10^{-3}$ \\
      & Std  & $1.38\times10^{-3}$ & $2.14\times10^{-5}$ & $2.33\times10^{-1}$ & $7.46\times10^{-1}$ & $3.95\times10^{-6}$ & $7.52\times10^{-15}$ & $2.06\times10^{-3}$ \\
\midrule
Matya & Mean & $0.00\times10^{0}$ & $0.00\times10^{0}$ & $4.69\times10^{-10}$ & $2.31\times10^{-1}$ & $3.15\times10^{-47}$ & $5.33\times10^{-20}$ & $3.46\times10^{-11}$ \\
      & Std  & $0.00\times10^{0}$ & $0.00\times10^{0}$ & $2.17\times10^{-9}$ & $2.07\times10^{-1}$ & $1.34\times10^{-46}$ & $2.85\times10^{-19}$ & $1.80\times10^{-10}$ \\
\midrule
Camel3 & Mean & $0.00\times10^{0}$ & $0.00\times10^{0}$ & $5.69\times10^{-11}$ & $2.87\times10^{-1}$ & $6.04\times10^{-75}$ & $4.17\times10^{-23}$ & $1.20\times10^{-15}$ \\
       & Std  & $0.00\times10^{0}$ & $0.00\times10^{0}$ & $1.33\times10^{-10}$ & $2.53\times10^{-1}$ & $3.24\times10^{-74}$ & $1.99\times10^{-22}$ & $3.35\times10^{-15}$ \\
\midrule
Exponential & Mean & $-1.00\times10^{0}$ & $-1.00\times10^{0}$ & $-1.00\times10^{0}$ & $-9.95\times10^{-1}$ & $-1.00\times10^{0}$ & $-1.00\times10^{0}$ & $-1.00\times10^{0}$ \\
            & Std  & $0.00\times10^{0}$ & $0.00\times10^{0}$ & $6.97\times10^{-13}$ & $6.08\times10^{-3}$ & $0.00\times10^{0}$ & $0.00\times10^{0}$ & $4.97\times10^{-17}$ \\
\midrule
DropWave & Mean & $-1.00\times10^{0}$ & $-1.00\times10^{0}$ & $-9.90\times10^{-1}$ & $-6.81\times10^{-1}$ & $-9.98\times10^{-1}$ & $-9.74\times10^{-1}$ & $-9.97\times10^{-1}$ \\
         & Std  & $0.00\times10^{0}$ & $0.00\times10^{0}$ & $2.20\times10^{-2}$ & $1.67\times10^{-1}$ & $1.14\times10^{-2}$ & $3.07\times10^{-2}$ & $1.24\times10^{-2}$ \\
\midrule
EggHolder & Mean & $-936.23$ & $-927.90$ & $-888.20$ & $-687.88$ & $-945.18$ & $-937.02$ & $-957.51$ \\
          & Std  & $31.63$ & $62.46$ & $81.13$ & $121.62$ & $36.86$ & $45.35$ & $4.72$ \\
\midrule
Himmelblau & Mean & $2.97\times10^{-8}$ & $7.40\times10^{-5}$ & $2.28\times10^{-9}$ & $1.00$ & $7.01\times10^{-5}$ & $6.00\times10^{-18}$ & $1.54\times10^{-1}$ \\
           & Std  & $1.30\times10^{-7}$ & $3.83\times10^{-4}$ & $4.28\times10^{-9}$ & $1.10$ & $8.43\times10^{-5}$ & $3.15\times10^{-17}$ & $1.48\times10^{-1}$ \\
\midrule
Levy13 & Mean & $6.25\times10^{-7}$ & $3.38\times10^{-7}$ & $3.79\times10^{-9}$ & $1.03$ & $4.26\times10^{-6}$ & $3.17\times10^{-19}$ & $1.30\times10^{-2}$ \\
       & Std  & $3.17\times10^{-6}$ & $7.61\times10^{-7}$ & $9.10\times10^{-9}$ & $1.52$ & $5.66\times10^{-6}$ & $1.39\times10^{-18}$ & $1.28\times10^{-2}$ \\
\midrule
Ackley01 & Mean & $4.44\times10^{-16}$ & $4.44\times10^{-16}$ & $8.83\times10^{-5}$ & $9.21$ & $4.44\times10^{-16}$ & $9.49\times10^{-12}$ & $5.76\times10^{-7}$ \\
         & Std  & $0.00\times10^{0}$ & $0.00\times10^{0}$ & $6.92\times10^{-5}$ & $3.97$ & $0.00\times10^{0}$ & $3.46\times10^{-11}$ & $1.82\times10^{-6}$ \\
\midrule
Griewank & Mean & $0.00\times10^{0}$ & $0.00\times10^{0}$ & $7.79\times10^{-3}$ & $2.59\times10^{-1}$ & $1.23\times10^{-3}$ & $1.24\times10^{-2}$ & $5.91\times10^{-3}$ \\
         & Std  & $0.00\times10^{0}$ & $0.00\times10^{0}$ & $6.99\times10^{-3}$ & $2.02\times10^{-1}$ & $2.76\times10^{-3}$ & $1.61\times10^{-2}$ & $8.47\times10^{-3}$ \\
\midrule
Michalewicz & Mean & $-1.801303$ & $-1.801303$ & $-1.747883$ & $-1.524660$ & $-1.801279$ & $-1.801303$ & $-1.604628$ \\
            & Std  & $6.84\times10^{-16}$ & $4.76\times10^{-10}$ & $2.00\times10^{-1}$ & $3.05\times10^{-1}$ & $2.70\times10^{-5}$ & $7.59\times10^{-16}$ & $2.91\times10^{-1}$ \\
\midrule
Qing & Mean & $7.78\times10^{-7}$ & $6.83\times10^{-4}$ & $1.34\times10^{-5}$ & $1.87\times10^{7}$ & $2.33\times10^{-6}$ & $2.05\times10^{-18}$ & $1.09\times10^{-2}$ \\
     & Std  & $2.92\times10^{-6}$ & $1.33\times10^{-3}$ & $3.64\times10^{-5}$ & $5.26\times10^{7}$ & $2.51\times10^{-6}$ & $8.47\times10^{-18}$ & $9.69\times10^{-3}$ \\
\midrule
Salomon & Mean & $0.00\times10^{0}$ & $0.00\times10^{0}$ & $2.43\times10^{-2}$ & $1.26$ & $3.33\times10^{-3}$ & $6.40\times10^{-2}$ & $1.45\times10^{-2}$ \\
        & Std  & $0.00\times10^{0}$ & $0.00\times10^{0}$ & $4.19\times10^{-2}$ & $6.94\times10^{-1}$ & $1.79\times10^{-2}$ & $4.72\times10^{-2}$ & $3.38\times10^{-2}$ \\
\midrule
Zimmerman & Mean & $0.35$ & $0.34$ & $846.11$ & $13281.45$ & $217.00$ & $0.40$ & $753.16$ \\
          & Std  & $0.35$ & $0.34$ & $608.19$ & $17820.68$ & $484.33$ & $0.34$ & $626.18$ \\
\midrule
Rana & Mean & $-499.36$ & $-499.96$ & $-498.85$ & $-405.87$ & $-500.27$ & $-499.54$ & $-500.10$ \\
     & Std  & $3.02$ & $0.66$ & $2.05$ & $51.42$ & $0.50$ & $1.88$ & $0.85$ \\
\midrule
Parsopoulos & Mean & $1.68\times10^{-4}$ & $5.97\times10^{-4}$ & $3.48\times10^{-10}$ & $1.68\times10^{-2}$ & $1.91\times10^{-6}$ & $2.78\times10^{-17}$ & $6.24\times10^{-4}$ \\
            & Std  & $3.77\times10^{-4}$ & $8.19\times10^{-4}$ & $1.08\times10^{-9}$ & $3.05\times10^{-2}$ & $6.51\times10^{-6}$ & $1.50\times10^{-16}$ & $1.02\times10^{-3}$ \\
\bottomrule
\end{tabular}
\end{table*}

Table \ref{tab:composition_results} displays the results for the CEC2017 composite functions with 100 dimensions. 40 independent trials have been conducted, with a population size of 50 and a maximum number of iterations of 1000. GEGO achieved superior results in eight out of the ten functions, further validating that the hybridization of GEO with GA significantly enhances the optimization capabilities of the original GEO algorithm.

\begin{table*}[]
\centering
\caption{Performance comparison on Composition Functions of CEC2017 (max\_it=1000, pop\_size=50, dim=100)}
\label{tab:composition_results}
\small
\setlength{\tabcolsep}{4pt} 
\begin{tabular}{llccccccc}
\toprule
Function & & GEGO & GEO & PSO & GA & GWO & JSA & SCA \\
\midrule
$CF1$ & Mean & $3286.00$ & $3359.29$ & $4.24\times10^{5}$ & $4330.64$ & $35891.46$ & $13651.29$ & $1.94\times10^{5}$ \\
         & Std  & $44.63$ & $48.59$ & $4.79\times10^{4}$ & $425.91$ & $11494.31$ & $4644.05$ & $12817.08$ \\
\midrule
$CF2$ & Mean & $2315.77$ & $2317.13$ & $2.08\times10^{4}$ & $2357.73$ & $2552.04$ & $2727.64$ & $11797.21$ \\
         & Std  & $18.75$ & $15.85$ & $5325.92$ & $18.02$ & $72.88$ & $111.52$ & $2575.08$ \\
\midrule
$CF3$ & Mean & $2797.56$ & $3248.38$ & $2.05\times10^{5}$ & $4577.79$ & $29488.87$ & $20023.16$ & $1.20\times10^{5}$ \\
         & Std  & $388.29$ & $1043.47$ & $6.49\times10^{4}$ & $655.39$ & $7030.03$ & $11449.19$ & $4328.49$ \\
\midrule
$CF4$ & Mean & $3072.76$ & $4307.67$ & $2.83\times10^{5}$ & $6651.94$ & $41606.15$ & $25690.22$ & $1.56\times10^{5}$ \\
         & Std  & $392.57$ & $1146.00$ & $8.34\times10^{4}$ & $1238.62$ & $8479.09$ & $7572.36$ & $7181.66$ \\
\midrule
$CF5$ & Mean & $3667.13$ & $3812.78$ & $6.27\times10^{4}$ & $3776.90$ & $5674.67$ & $4529.59$ & $20990.58$ \\
         & Std  & $75.73$ & $51.74$ & $1.90\times10^{4}$ & $117.71$ & $750.35$ & $277.99$ & $2531.21$ \\
\midrule
$CF6$ & Mean & $4804.57$ & $4934.42$ & $3.01\times10^{4}$ & $6076.91$ & $7946.87$ & $6200.89$ & $64668.72$ \\
         & Std  & $337.82$ & $332.02$ & $11921.00$ & $226.72$ & $795.20$ & $173.76$ & $11409.76$ \\
\midrule
$CF7$ & Mean & $3516.47$ & $3549.59$ & $5022.08$ & $3634.71$ & $3579.54$ & $3685.26$ & $6886.35$ \\
         & Std  & $40.57$ & $38.39$ & $654.26$ & $126.69$ & $89.37$ & $121.29$ & $382.97$ \\
\midrule
$CF8$ & Mean & $2910.77$ & $2983.32$ & $16762.54$ & $3284.21$ & $3767.52$ & $3469.76$ & $13591.13$ \\
         & Std  & $90.43$ & $66.17$ & $3507.38$ & $33.95$ & $176.45$ & $100.75$ & $1443.80$ \\
\midrule
$CF9$ & Mean & $6.06\times10^{7}$ & $2.69\times10^{8}$ & $5.29\times10^{15}$ & $8.39\times10^{7}$ & $2.02\times10^{9}$ & $2.53\times10^{6}$ & $4.91\times10^{13}$ \\
         & Std  & $3.37\times10^{7}$ & $1.05\times10^{8}$ & $1.40\times10^{16}$ & $2.65\times10^{8}$ & $1.26\times10^{9}$ & $2.71\times10^{6}$ & $5.08\times10^{13}$ \\
\midrule
$CF10$ & Mean & $2.52\times10^{8}$ & $9.10\times10^{8}$ & $2.15\times10^{15}$ & $4.24\times10^{7}$ & $1.14\times10^{10}$ & $2.32\times10^{8}$ & $3.65\times10^{13}$ \\
         & Std  & $9.71\times10^{7}$ & $2.64\times10^{8}$ & $3.36\times10^{15}$ & $9.45\times10^{7}$ & $5.81\times10^{9}$ & $6.66\times10^{7}$ & $5.30\times10^{13}$ \\
    
\bottomrule
\end{tabular}
\end{table*}

\begin{table*}[]
\centering
\caption{Results of benchmark functions (max\_it=100, pop\_size=20)}
\label{tab:math_simple_20_2}
\small
\setlength{\tabcolsep}{5pt}
\begin{tabular}{llccc}
\toprule
Function & & GEGO & L-SHADE & CMA-ES \\
\midrule
Beale & Mean & $5.16\times10^{-4}$ & $2.54\times10^{-2}$ & $7.30\times10^{-1}$ \\
      & Std  & $1.38\times10^{-3}$ & $1.37\times10^{-1}$ & $6.99\times10^{-1}$ \\
\midrule
Matya & Mean & $0.00\times10^{0}$ & $2.22\times10^{-16}$ & $1.40\times10^{-2}$ \\
      & Std  & $0.00\times10^{0}$ & $4.81\times10^{-16}$ & $3.16\times10^{-2}$ \\
\midrule
Camel3 & Mean & $0.00\times10^{0}$ & $1.57\times10^{-19}$ & $5.29\times10^{-2}$ \\
       & Std  & $0.00\times10^{0}$ & $3.14\times10^{-19}$ & $1.11\times10^{-1}$ \\
\midrule
Exponential & Mean & $-1.00\times10^{0}$ & $-1.00\times10^{0}$ & $-9.99\times10^{-1}$ \\
            & Std  & $0.00\times10^{0}$ & $0.00\times10^{0}$ & $2.43\times10^{-3}$ \\
\midrule
DropWave & Mean & $-1.00\times10^{0}$ & $-9.97\times10^{-1}$ & $-9.14\times10^{-1}$ \\
         & Std  & $0.00\times10^{0}$ & $1.17\times10^{-2}$ & $5.39\times10^{-2}$ \\
\midrule
EggHolder & Mean & $-9.36\times10^2$ & $-9.31\times10^{2}$ & $-5.87\times10^{2}$ \\
          & Std  & $31.63$ & $4.80\times10^{1}$ & $1.37\times10^{2}$ \\
\midrule
Himmelblau & Mean & $2.97\times10^{-8}$ & $9.40\times10^{-8}$ & $5.89\times10^{0}$ \\
           & Std  & $1.30\times10^{-7}$ & $4.21\times10^{-7}$ & $6.64\times10^{0}$ \\
\midrule
Levy13 & Mean & $6.25\times10^{-7}$ & $2.62\times10^{-17}$ & $4.47\times10^{-1}$ \\
       & Std  & $3.17\times10^{-6}$ & $1.06\times10^{-16}$ & $5.85\times10^{-1}$ \\
\midrule
Ackley01 & Mean & $4.44\times10^{-16}$ & $1.94\times10^{-9}$ & $3.72\times10^{0}$ \\
         & Std  & $0.00\times10^{0}$ & $2.88\times10^{-9}$ & $1.81\times10^{0}$ \\
\midrule
Griewank & Mean & $0.00\times10^{0}$ & $6.39\times10^{-4}$ & $7.14\times10^{-2}$ \\
         & Std  & $0.00\times10^{0}$ & $1.84\times10^{-3}$ & $5.45\times10^{-2}$ \\
\midrule
Michalewicz & Mean & $-1.801303$ & $-1.801303$ & $-1.637323$ \\
            & Std  & $6.84\times10^{-16}$ & $2.66\times10^{-16}$ & $1.93\times10^{-1}$ \\
\midrule
Qing & Mean & $7.78\times10^{-7}$ & $3.76\times10^{-8}$ & $1.05\times10^{5}$ \\
     & Std  & $2.92\times10^{-6}$ & $1.19\times10^{-7}$ & $3.99\times10^{5}$ \\
\midrule
Salomon & Mean & $0.00\times10^{0}$ & $1.91\times10^{-2}$ & $4.56\times10^{-1}$ \\
        & Std  & $0.00\times10^{0}$ & $3.30\times10^{-2}$ & $2.63\times10^{-1}$ \\
\midrule
Zimmerman & Mean & $3.50\times10^{-1}$ & $1.87\times10^{-1}$ & $4.21\times10^{4}$ \\
          & Std  & $0.35$ & $3.09\times10^{-1}$ & $3.56\times10^{4}$ \\
\midrule
Rana & Mean & $-499.36$ & $-4.99\times10^{2}$ & $-3.73\times10^{2}$ \\
     & Std  & $3.02$ & $2.71\times10^{0}$ & $6.32\times10^{1}$ \\
\midrule
Parsopoulos & Mean & $1.68\times10^{-4}$ & $5.52\times10^{-8}$ & $1.96\times10^{-2}$ \\
            & Std  & $3.77\times10^{-4}$ & $2.61\times10^{-7}$ & $2.17\times10^{-2}$ \\
\bottomrule
\end{tabular}
\end{table*}
For further comparison, GEGO was compared with newer and state of the art evolutionary algorithms, such as Linear Population Size Reduction Success-History Adaptation Differential Evolution (L-SHADE) algorithm \cite{lshade} and Covariance Matrix Adaptation Evolution Strategy (CMA-ES) \cite{cmes}. Table \ref{tab:math_simple_20_2} presents the results for the unimodal and multimodal functions, where GEGO outperformed in most of the functions. 

\begin{table*}[]
\centering
\caption{Performance comparison on Composition Functions of CEC2017 (max\_it=1000, pop\_size=50, dim=100)}
\label{tab:composition_results_new}
\small
\setlength{\tabcolsep}{6pt}
\begin{tabular}{llccc}
\toprule
Function & & GEGO & L-SHADE & CMA-ES \\
\midrule
$CF1$ & Mean & $3286.00$ & $3226.06$ & $2.47\times10^{5}$ \\
& Std  & $44.63$ & $221.97$ & $13986.37$ \\
\midrule
$CF2$ & Mean & $2315.77$ & $2377.71$ & $28167.99$ \\
& Std  & $18.75$ & $16.44$ & $1715.23$ \\
\midrule
$CF3$ & Mean & $2797.56$ & $2519.99$ & $1.30\times10^{5}$ \\
& Std  & $388.29$ & $318.45$ & $5173.99$ \\
\midrule
$CF4$ & Mean & $3072.76$ & $2526.92$ & $1.82\times10^{5}$ \\
& Std  & $392.57$ & $164.88$ & $8054.48$ \\
\midrule
$CF5$ & Mean & $3667.13$ & $3361.00$ & $35323.44$ \\
& Std  & $75.73$ & $44.26$ & $4223.40$ \\
\midrule
$CF6$ & Mean & $4804.57$ & $5719.30$ & $2.91\times10^{5}$ \\
& Std  & $337.82$ & $384.01$ & $37921.93$ \\
\midrule
$CF7$ & Mean & $3516.47$ & $3310.64$ & $13386.17$ \\
& Std  & $40.57$ & $43.56$ & $1049.54$ \\
\midrule
$CF8$ & Mean & $2910.77$ & $2819.07$ & $24496.28$ \\
& Std  & $90.43$ & $106.07$ & $2061.41$ \\
\midrule
$CF9$ & Mean & $6.06\times10^{7}$ & $1.55\times10^{4}$ & $4.65\times10^{15}$ \\
& Std  & $3.37\times10^{7}$ & $3755.38$ & $3.50\times10^{15}$ \\
\midrule
$CF10$ & Mean & $2.52\times10^{8}$ & $6.59\times10^{4}$ & $1.10\times10^{15}$ \\
& Std  & $9.71\times10^{7}$ & $26378.94$ & $7.22\times10^{14}$ \\
\bottomrule
\end{tabular}
\end{table*}

Similar to the previous experiment, 40 independent trials were conducted with a population size of 50 and a maximum of 1000 iterations for the CEC2017 composite functions with 100 dimensions. The results are reported in Table \ref{tab:composition_results_new}. GEGO outperformed CMA-ES across all ten composite functions, while its performance relative to L-SHADE was more limited, with GEGO achieving better results in two out of ten functions, reflecting the strong optimization capability of L-SHADE on large-scale composite benchmarks.

\subsection{ANN Hyperparameter Tuning}
A significant application for population-based optimization algorithms is the hyperparameter tuning of neural networks. Various hyperparameters profoundly influence the performance of Artificial Neural Network (ANN) models; consequently, their optimization is a critical stage in the training pipeline. 

To evaluate the algorithms in this context, the objective function was defined as the test accuracy, which the algorithms aimed to maximize. Computationally, this was implemented by minimizing the negative test accuracy. The experiments were conducted using the MNIST dataset, which contains 70,000 images (60,000 for training and 10,000 for testing). To accommodate hardware constraints while maintaining a rigorous comparison, we evaluated the performance of GEGO against its constituent algorithms, GEO and GA.

To ensure a fair comparison, a fixed random seed was used for data partitioning, and identical search space boundaries were applied to all algorithms. The implementation was developed in Python using the TensorFlow framework. It is important to clarify the distinct roles of metaheuristic optimization and gradient-based learning in this study. The proposed GEGO, along with GEO and GA, is used exclusively for hyperparameter optimization, including the selection of network architecture parameters and training-related hyperparameters. Once a candidate hyperparameter configuration is generated by a metaheuristic algorithm, the corresponding ANN is trained using the Adam optimizer, which is employed solely for updating network weights via backpropagation. Adam is not involved in the hyperparameter search process. This separation ensures that all metaheuristic algorithms are evaluated fairly using an identical gradient-based training procedure while focusing their role on global hyperparameter optimization. 

The search space for the particles are continuous, and it is then mapped to the discrete values shown in the table \ref{tab:ann_structure}. The particles have a dimension of ten. The first eight values represent pairs of the number of neurons and the dropout rate of each layer, respectively. 
The last two values show the batch size and learning rate. If the number of neurons in that layer is less than the number in the output layer, which in the case of MNIST is ten, then that layer and the following layers are excluded. This approach allows the ANN structure to have fewer layers than four. The real value is floored to determine the number of neurons for each layer, the index for the dropout rate, and the batch size. The learning rate is left as a real value since it does not need to be an integer. 

It is important to clarify the scope of the optimization task considered in this study. Although GEGO is used to optimize certain architecture related hyperparameters such as the number of hidden layers and the number of neurons per layer, the proposed approach does not constitute a full neural architecture search framework. The search space is restricted to a predefined family of fully connected feedforward networks with limited depth, and no optimization of network topology, connectivity patterns, or computational graph structure is performed. All candidate models share the same basic architectural template, and GEGO is employed solely to select hyperparameter values within this fixed design space. Accordingly, the proposed method is more appropriately characterized as hyperparameter optimization with architectural parameters rather than neural architecture search in the broader sense.

The search space for the particles is detailed in Table \ref{tab:ann_structure}, which includes the number of neurons per layer, the presence and rate of dropout layers, the learning rate, and the batch size.

\begin{table}[]
\centering
\caption{ANN STRUCRURE}
\label{tab:ann_structure}
\resizebox{\columnwidth}{!}{
\begin{tabular}{|c|c|}
\hline
Number of layers & {1,2,3,4} \\ 
\hline
First layer's number of neurons & [128:512] \\
\hline
Second layer's number of neurons  & [64:512] \\
\hline
Third layer's number of neurons  & [32:512]\\
\hline
Forth layer's number of neurons  & [16:512]\\
\hline
Dropout rate & {0,0.1,0.5}\\
\hline
Batch size & {32,64,96,128,160,192,224,256}\\
\hline
Learning rate & [1e-2,1e-4]\\
\hline
\end{tabular}
}
\end{table}

Initial experiments included regularization techniques and various activation functions within the search space. However, preliminary results on the MNIST dataset indicated that these often led to poor convergence, with accuracy stagnating around 10\%. This increased the computational overhead without providing significant insights. Consequently, the final experiments utilized the ReLU activation function for internal layers and Softmax for the output layer, with no additional regularization. Training was conducted with a maximum of 50 epochs and an early stopping criterion with a patience of 5 iterations. The training parameters are summarized in Table \ref{tab:training_param}.
\begin{table}[]
\centering
\caption{TRAINING MODEL PARAMETERS}
\label{tab:training_param}
\resizebox{\columnwidth}{!}{
\begin{tabular}{|c|c|}
\hline
\textbf{Early stopping parameters}  \\ 
\hline
monitor & val\_loss \\
\hline
patience & 5 \\
\hline
restore\_best\_weights  & true\\
\hline
min\_delta  & 0.005\\
\hline
Drop\_out rate & {0,0.1,0.5}\\
\hline
\textbf{Training parameters} & \\
\hline
epochs & 50\\
\hline
optimizer & Adam\\
\hline
loss & sparse\_categorical\_crossentropy\\
\hline
\end{tabular}
}
\end{table}

While mathematical benchmarks allow for large populations (30+ particles) and high iteration counts (1,000+) due to their near-instantaneous evaluation, training an ANN is significantly more computationally intensive. The training time for a single ANN varies based on batch size and network architecture; thus, the population size and iterations must be carefully managed. Table \ref{tab:ann_results} presents the results for GA, GEO, and GEGO using a population of 10 particles over 15 iterations across ten independent trials. 
The choice of a relatively small population size and limited number of iterations was driven by the high computational cost of training neural networks compared to mathematical benchmark functions. Each fitness evaluation requires training and evaluating an ANN model, and therefore the optimizer depth must be carefully balanced against available computational resources. The selected configuration allows meaningful comparison between optimization algorithms while maintaining reproducibility and reasonable execution time.

Early stopping based on validation performance further supports efficient training by preventing unnecessary epochs once convergence behavior stabilizes. This design reflects realistic scenarios in which practitioners must optimize model configurations under constrained budgets, and it highlights the practical applicability of metaheuristic optimization in real-world machine learning pipelines.
\begin{table}[ht]
\centering
\caption{performance comparison of geo, gego, and ga}
\label{tab:ann_results}
\begin{tabular}{lccc}
\hline
Metric & GEO & GEGO & GA \\
\hline
Mean (\%) & 97.46 & 97.62 & 97.12 \\
Std (\%) & 0.05 & 0.16 & 0.14 \\
Max Test Accuracy (\%) & 97.60 & 97.90 & 97.31 \\
Max Train Accuracy (\%) & 97.85 & 97.95 & 97.76 \\
Loss & 0.07163 & 0.06575 & 0.07242 \\
\hline
\end{tabular}
\end{table}

GEGO demonstrated superior average performance compared to both GEO and GA. While GEO exhibited the lowest standard deviation, this was indicative of a tendency to converge prematurely to a local maximum (approximately 97.4\% accuracy). In contrast, GEGO successfully navigated away from these local optima. The highest overall performance was achieved by GEGO, reaching a training accuracy of 97.95\% and a test accuracy of 97.90\% with a loss of 0.06. The optimal parameters identified by each algorithm are presented in Table \ref{tab:ann_best}.

\begin{table}[ht]
\centering
\caption{hyperparameters found for the ann}
\label{tab:ann_best}
\resizebox{\columnwidth}{!}{
\begin{tabular}{lcccc}
\hline
Algorithm & Number of Neurons & Dropout Rate & Batch Size & Learning Rate \\
\hline
GA   & [356, 388, 413, 80]  & [0.1, 0, 0, 0]   & 224 & 0.01 \\
GEO  & [512, 135, 504, 220] & [0.5, 0.5, 0, 0] & 32  & 0.01 \\
GEGO & [378, 191, 220, 106] & [0, 0.1, 0, 0]   & 224 & 0.01 \\
\hline
\end{tabular}
}
\end{table}

\section{Conclusion}
In this study, we introduced the Golden Eagle Genetic Optimization (GEGO) algorithm, a hybrid metaheuristic that integrates the search dynamics of Golden Eagle Optimization (GEO) with the evolutionary operators of Genetic Algorithms (GA). The proposed framework was evaluated on a diverse set of mathematical benchmark functions, including the CEC2017 composite functions, where GEGO consistently demonstrated improved solution quality and robustness compared to its constituent algorithms and several commonly used metaheuristic methods.

Beyond mathematical benchmarks, GEGO was applied to the practical task of hyperparameter tuning for Artificial Neural Networks (ANNs). The experimental results indicate that GEGO effectively explores complex and high dimensional hyperparameter spaces and achieves more stable convergence than both GEO and GA. In particular, the hybrid design enables GEGO to escape local optima that frequently limit the performance of the standard GEO algorithm. By combining the directed swarm intelligence of GEO with the diversity preserving mechanisms of GA, GEGO provides an effective and computationally efficient optimization framework for machine learning applications.

An important practical motivation for this study is the prevalence of limited computational resources in many academic institutions and small research groups, where access to large scale GPU clusters is often restricted. In such settings, exhaustive hyperparameter search and large scale optimization are frequently infeasible. The proposed GEGO framework is particularly well suited to these scenarios, as it is designed to deliver robust optimization performance under constrained computational budgets while maintaining modest population sizes and iteration counts.

Despite the encouraging results, the present study is limited to controlled experimental settings and moderate scale neural network architectures due to computational constraints. Future work will investigate the scalability of GEGO on larger datasets and more complex deep learning models, as well as conduct more extensive statistical significance analysis using a greater number of independent runs when additional computational resources become available.

While GEGO demonstrates competitive performance across a range of benchmark functions and consistently outperforms several baseline optimizers, its performance relative to advanced differential evolution variants such as L-SHADE is more mixed, particularly on large scale composite benchmark problems. A more extensive evaluation under larger population sizes and higher function evaluation budgets is therefore identified as an important direction for future work.



\section*{Institutional review} Not Applicable

\section*{Informed consent} Not Applicable

\section*{Data availability} The code is available on a reasonable request to the first author Amaras Nazarians.


\section*{Acknowledgments} Not Applicable

\section*{Conflicts of interest} The authors declare no conflicts of interest.

\printbibliography

\end{document}